\documentclass[conference,a4paper]{IEEEtran}
\makeatletter
\newcommand{\linebreakand}{%
  \end{@IEEEauthorhalign}
  \hfill\mbox{}\par
  \mbox{}\hfill\begin{@IEEEauthorhalign}
}
\makeatother

\IEEEoverridecommandlockouts

\usepackage{cite}
\usepackage{amsmath,amssymb,amsfonts}
\usepackage{graphicx}
\usepackage{textcomp}
\usepackage{xcolor}
\usepackage{booktabs}
\usepackage{multirow}
\usepackage{array}
\usepackage{flafter}
\usepackage[
    a4paper,
    top=19.1mm,
    bottom=19.1mm,
    left=14.3mm,
    right=14.3mm
]{geometry}

\def\BibTeX{{\rm B\kern-.05em{\sc i\kern-.025em b}\kern-.08em
    T\kern-.1667em\lower.7ex\hbox{E}\kern-.125emX}}

\begin{document}

\flushbottom

\title{Accuracy- and Real-Time-Aware 4D Radar Preprocessing for Autonomous Driving Perception Systems}

\author{
\IEEEauthorblockN{1\textsuperscript{st} Woo-Jin Jung}
\IEEEauthorblockA{\textit{Cho Chun Shik Graduate School of Mobility} \\
\textit{Korea Advanced Institute of Science and Technology (KAIST)}\\
Daejeon 34051, Republic of Korea}
\and
\IEEEauthorblockN{2\textsuperscript{nd} Dong-Hee Paek}
\IEEEauthorblockA{\textit{Mechanical Engineering Research Institute} \\
\textit{Korea Advanced Institute of Science and Technology (KAIST)}\\
Daejeon 34141, Republic of Korea}
\linebreakand
\IEEEauthorblockN{3\textsuperscript{rd} Jeong-Su Park}
\IEEEauthorblockA{\textit{Autonomous Driving Perception Technology Vanguard Team} \\
\textit{Hyundai Motor Company}\\
Seongnam-si 13529, Republic of Korea}
\and
\IEEEauthorblockN{4\textsuperscript{th} Seung-Hyun Kong\textsuperscript{*}\thanks{*Corresponding author}}
\IEEEauthorblockA{\textit{Cho Chun Shik Graduate School of Mobility} \\
\textit{Korea Advanced Institute of Science and Technology (KAIST)}\\
Daejeon 34051, Republic of Korea\\
skong@kaist.ac.kr}
}

\maketitle

% Preserve the reference order used in the source Word manuscript.
\nocite{paek2022kradar,jung2026lidar,cho2019enhancing,roldan2024deep,kong2014fast,
sun2022lidar,moon2024adaptive,kong2021enhanced,jalil2016analysis,tan2022multiframe,
geiger2012kitti,caesar2020nuscenes,wong2019netscore,richards2010principles,
izacard2019data,liu2023smurf,chen2017tutorial,bundy1984dog,zhang2025dual,
kong2025rtnh,musiat2024radarpillars,xu2021rpfa,lang2019pointpillars,
leitgeb2026rade,haitman2025doppdrive,wang2025dadan,bentley1975multidimensional}

\begin{abstract}
4D radar has emerged as a promising next-generation sensor for improving the robustness of autonomous driving perception systems because of its stable sensing capability under adverse weather conditions. However, deploying 4D radar in embedded environments with limited hardware resources requires radar-representation preprocessing that jointly considers perception accuracy, real-time performance, and computational complexity. This paper proposes a preprocessing framework for 4D-radar-based 3D object detection. First, Percentile-based 3D Shape Preservation (P3DP) extracts point clouds from radar tensors while preserving object-shape information and suppressing noise and false alarms. Second, Multi-frame-based Noise Point Discrimination using Kernel Density Estimation (MF-KDE) improves the density and reliability of sparse radar point clouds. Finally, Embedded \& NetScore (ENS) evaluates suitability for embedded deployment by jointly considering accuracy, real-time performance, adverse-weather robustness, and model complexity.
\end{abstract}


\section{Introduction}

4D radar measures the three-dimensional position of an object by providing elevation information in addition to the range, azimuth, and Doppler information available from conventional automotive radar. Because 4D radar uses electromagnetic waves with longer wavelengths than those used by cameras or LiDAR, it provides relatively stable measurements under adverse weather conditions such as rain, fog, and snow \cite{paek2022kradar,jung2026lidar,cho2019enhancing}. It can also directly estimate object velocity through the Doppler effect. These properties have made 4D radar a promising next-generation sensor for improving the robustness of autonomous driving perception systems \cite{jung2026lidar}.

To use 4D radar in a deep neural network (DNN)-based autonomous driving perception system, radar measurements must be converted into an appropriate input representation \cite{paek2022kradar,jung2026lidar}. Two representative radar representations are the 4D radar tensor and the point cloud. A 4D radar tensor expresses radar measurements on a grid and contains rich information along the range, azimuth, elevation, and Doppler axes. A point cloud is a set of selected reflections represented by 3D coordinates and radar features after noise such as clutter and multipath reflections has been removed from the tensor. Recent studies have increasingly used raw measurements such as 4D radar tensors as input because they minimize information loss and retain more radar information than point clouds \cite{paek2022kradar,jung2026lidar,cho2019enhancing,roldan2024deep,kong2014fast}. However, the large size of a tensor imposes substantial computational and data-processing burdens when it is used directly as real-time DNN input in a vehicle. In contrast, a point cloud is much smaller and therefore more suitable for autonomous driving perception systems \cite{paek2022kradar,sun2022lidar}.

The performance of a point-cloud representation nevertheless depends on how it is generated. Tensor preprocessing extracts a point cloud suitable for DNN input from a 4D radar tensor. Constant False Alarm Rate (CFAR) and percentile-based methods are commonly used for this purpose \cite{paek2022kradar,moon2024adaptive}. CFAR detects cells whose reflected power is high relative to that of surrounding cells, whereas a percentile-based method selects a fraction of the cells with the highest power values in a tensor. These methods can extract strong reflections in relatively simple backgrounds. In an autonomous driving environment containing complex road structures, nearby vehicles, clutter, and multipath reflections, however, they have difficulty generating a point cloud that both preserves object measurements and suppresses noise and false alarms \cite{jung2026lidar}. Radar point clouds are also generally sparse and do not sufficiently represent object surfaces and boundaries, limiting the ability of a DNN to learn object position, size, and orientation reliably \cite{paek2022kradar,jung2026lidar}. Effective application of 4D radar to autonomous driving perception therefore requires both tensor preprocessing that suppresses noise and false alarms while preserving object-shape information and point-cloud preprocessing that improves the density and reliability of sparse measurements \cite{kong2021enhanced}.

\begin{figure}[!htbp]
    \centering
    \includegraphics[width=\columnwidth]{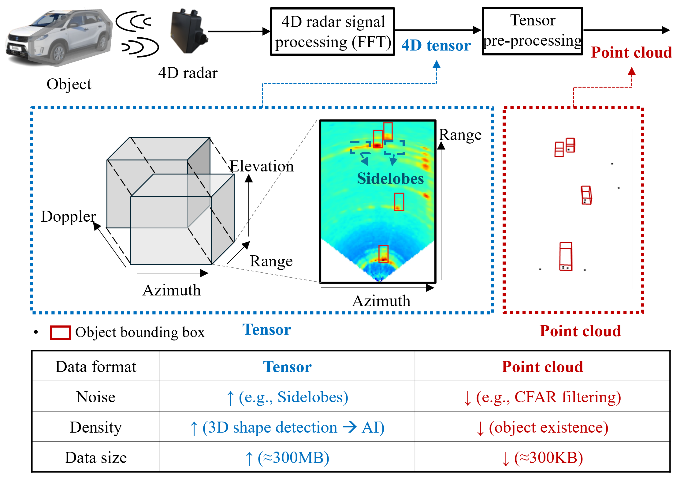}
    \caption{Data representations of 4D radar. A tensor provides richer spatial information but is computationally expensive, whereas a point cloud is lightweight but sparse and less effective for representing object shape.}
    \label{fig:radar_representations}
\end{figure}

Figure~\ref{fig:radar_representations} compares the information content and computational characteristics of the two radar representations.

This paper proposes tensor- and point-cloud-preprocessing methods for 4D-radar-based 3D object detection. For tensor preprocessing, we propose Percentile-based 3D Shape Preservation (P3DP), which extracts a point cloud suitable for DNN input while preserving the 3D shape of an object. P3DP extracts object-relevant points and suppresses noise and false alarms. For point-cloud preprocessing, we propose Multi-frame-based Noise Point Discrimination using Kernel Density Estimation (MF-KDE). MF-KDE increases point density through multi-frame alignment and adds a kernel-density-estimation (KDE)-based density feature that helps a DNN distinguish object-related points from noise. Finally, to evaluate suitability for practical embedded autonomous driving perception systems, we propose Embedded \& NetScore (ENS), which jointly considers object-detection accuracy, real-time performance, adverse-weather robustness, and model complexity. This approach is consistent with prior work emphasizing that autonomous driving systems should be evaluated in terms of data-use efficiency, training stability, and computational efficiency as well as model performance \cite{kong2021enhanced}.

\section{Related Work}

This section reviews representative radar-preprocessing methods, trends in 4D-radar-based object detection, and existing evaluation metrics for embedded environments.

\subsection{Existing Preprocessing Methods}

\begin{figure}[!htbp]
    \centering
    \includegraphics[width=\columnwidth]{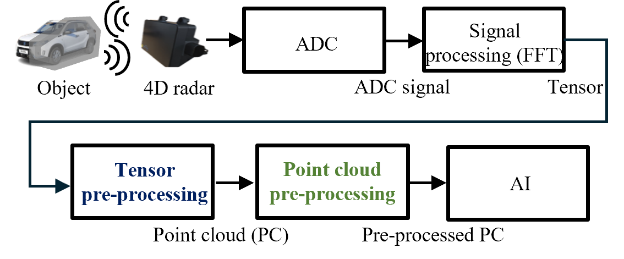}
    \caption{Overview of the 4D radar perception system. ADC denotes analog-to-digital converter. Tensor preprocessing and point-cloud preprocessing generate radar representations that are subsequently used as inputs to the DNN-based object-detection network.}
    \label{fig:perception_overview}
\end{figure}

Radar preprocessing constructs a point cloud that enables a DNN to learn an object's position, size, orientation, and shape from radar measurements. Existing methods can be divided into tensor preprocessing, which extracts a point cloud from a tensor, and point-cloud preprocessing, which improves the density and reliability of a sparse point cloud. Figure~\ref{fig:perception_overview} shows where these operations occur in a 4D radar perception system.

\subsubsection{Tensor Preprocessing}

The central objective of tensor preprocessing is to extract points that are useful for object detection without losing the object's 3D shape. Representative methods include CFAR \cite{tan2022multiframe} and percentile-based point extraction \cite{paek2022kradar}. CFAR detects cells with high reflected power relative to surrounding cells, but its independent selection of points makes it difficult to preserve an object's spatial structure. Percentile-based preprocessing simply and efficiently selects the top $K\%$ of cells by power. However, it may also select high-power noise, producing a tradeoff between false alarms and missed detections \cite{paek2022kradar}.

Because point extraction can discard valid measurements, recent studies directly use tensors or compressed tensor representations. For example, the Tensor Projection Module (TPM) of RADE-Net \cite{leitgeb2026rade} projects the Doppler and elevation axes of a 4D tensor by their maximum values and combines the projections, reducing the data size from approximately 260~MB to 21~MB. The representation nevertheless remains much larger than a point cloud, and TPM was not optimized for real-time processing, imposing a processing burden in embedded environments.

\subsubsection{Point-Cloud Preprocessing}

Radar point clouds generally have lower resolution and more noise than camera or LiDAR data. The radar's internal detection and filtering process also leaves few measurements for each object, so object surfaces and boundaries are not represented continuously and the 3D shape of an object is not sufficiently preserved. Existing methods address this sparsity through multi-frame alignment, which transforms point clouds from previous frames into the coordinate system of the current frame and accumulates measurements from multiple time steps to increase point density \cite{tan2022multiframe}. Simple frame accumulation, however, increases not only object-related points but also noise points, changes in the positions of moving objects, and inaccurate points caused by alignment errors. DoppDrive \cite{haitman2025doppdrive} corrects previous points in the radial direction to reduce the dispersion caused by moving objects. Because it does not directly correct tangential motion, residual position errors can remain for objects with substantial lateral motion; removing historical points with large predicted tangential errors can also reduce temporal information.

\subsection{4D-Radar-Based Object Detection}

Research on 4D radar object detection has evolved toward compensating for the inherent sparsity and noise of radar data. RPFA-Net \cite{xu2021rpfa} learns spatial relations between pillars through self-attention, and MF-Net \cite{tan2022multiframe} uses multiple frames and a spatiotemporal encoder. SMURF \cite{liu2023smurf} addresses sparsity through KDE-based density features and multi-representation fusion. DADAN \cite{wang2025dadan} uses a density-aware architecture to mitigate the decreasing point density with range and the difficulty of distinguishing objects from noise. RadarPillar-Net \cite{musiat2024radarpillars} extracts position, Doppler, and radar cross-section (RCS) features separately to use radar-specific information effectively.

\subsection{Evaluation Metrics for Embedded Environments}

Object detection is commonly evaluated using Average Precision (AP) \cite{geiger2012kitti}. Bird's-eye-view AP (BEV AP) measures detection performance in the bird's-eye view, whereas 3D AP measures detection performance in three-dimensional space. Practical deployment in an in-vehicle embedded environment, however, requires consideration of real-time performance, computational cost, and model size in addition to accuracy. The nuScenes Detection Score (NDS) \cite{caesar2020nuscenes} is a representative metric that combines AP with errors in object position, size, orientation, and velocity, but it does not directly evaluate computational cost or model complexity. NetScore \cite{wong2019netscore} was proposed to assess deep-learning efficiency by combining model accuracy, parameter count, and computational complexity. Because NetScore uses only AP-based accuracy, it does not fully reflect 3D detection quality or robustness to adverse weather, both of which are important in autonomous driving.

\section{Proposed Preprocessing Methods}

This section presents the proposed tensor- and point-cloud-preprocessing methods for 4D-radar-based 3D object detection. Both methods suppress noise in radar measurements and organize the measurements so that a DNN can effectively learn object position, size, orientation, and shape.

\subsection{4D Radar Tensor Preprocessing: P3DP}

\begin{figure}[!htbp]
    \centering
    \includegraphics[width=\columnwidth]{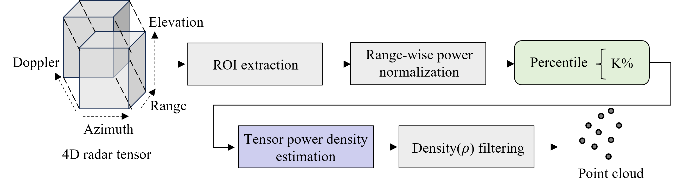}
    \caption{Overall framework of P3DP.}
    \label{fig:p3dp_pipeline}
\end{figure}

A 4D radar tensor preserves an object's spatial information without the loss caused by point extraction, but it contains noise such as clutter and is much larger than a point cloud. Direct use as DNN input in an autonomous driving environment can therefore limit both real-time performance and accuracy. We propose Percentile-based 3D Shape Preservation (P3DP) to reduce the data volume and remove noise while retaining useful spatial information in the tensor. As shown in Fig.~\ref{fig:p3dp_pipeline}, P3DP comprises five stages: region-of-interest (ROI) extraction, range-wise power normalization, percentile-based point extraction, tensor power-density estimation, and density filtering.

\subsubsection{ROI Extraction}

We first define a region of interest (ROI) within the 4D radar tensor for object detection. Using the entire tensor increases the computational cost and can introduce noise from regions that are not directly relevant to perception in a driving environment. We therefore use the following ROI, which is the same as the configuration provided by the dataset and reflects the forward-driving environment:

\begin{equation}
    0 < x < 73, \qquad -16 < y < 16, \qquad -2 < z < 6,
    \label{eq:roi}
\end{equation}
where $x$, $y$, and $z$ denote the longitudinal, lateral, and vertical positions in meters, respectively.

\subsubsection{Range-Wise Power Normalization}

Radar reflected power decreases as the distance between an object and the sensor increases. Consequently, the same object can produce a strong power response at short range and a weak response at long range \cite{liu2023smurf}. If CFAR or percentile-based point extraction is performed without compensating for range, relatively few points may be extracted from distant objects, causing a loss of their shape information.

P3DP mitigates this problem through range-wise power normalization. Specifically, the power of each tensor cell is multiplied by the square of its range:

\begin{equation}
    P_{\mathrm{norm}}(r,\theta,\phi,v)
    = r^{2} P(r,\theta,\phi,v),
    \label{eq:range_normalization}
\end{equation}
where $r$ is the range of the tensor cell, $\theta$ is the azimuth angle, $\phi$ is the elevation angle, $v$ is the Doppler velocity, $P(r,\theta,\phi,v)$ is the original radar power, and $P_{\mathrm{norm}}(r,\theta,\phi,v)$ is the range-normalized radar power. This normalization compensates for the weaker power response of distant objects, allowing objects at different ranges to be selected more evenly during percentile-based point extraction.

\subsubsection{Percentile-Based Point Extraction}

After range-wise power normalization, P3DP extracts as candidate points the top $K\%$ of cells with the highest normalized power. This procedure exploits the tendency of radar signals reflected by objects to have greater power than noise. We set $K=10\%$ in the experiments.

\subsubsection{Tensor Power-Density Estimation}

The central idea of P3DP is to consider the power distribution around each cell rather than selecting a point solely from the power of that cell. Conventional percentile-based filtering selects high-power cells, but road structures, background reflections, and sidelobes around objects can also generate strong power responses in autonomous driving environments \cite{richards2010principles,izacard2019data}. A criterion based only on power may therefore select noise points unrelated to object shape.

\begin{figure}[!htbp]
    \centering
    \begin{minipage}[c]{0.25\columnwidth}
        \centering
        \includegraphics[width=\linewidth]{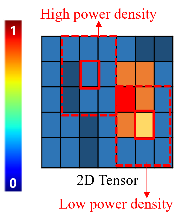}\\[-2pt]
        \footnotesize (a)
    \end{minipage}
    \hfill
    \begin{minipage}[c]{0.70\columnwidth}
        \centering
        \includegraphics[width=\linewidth]{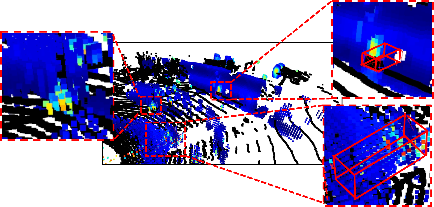}\\[-2pt]
        \footnotesize (b)
    \end{minipage}
    \caption{(a) Tensor power-density estimation and (b) visualization of the power density. Black points denote LiDAR measurements, and colored points denote percentile-extracted radar points. Red indicates high power in (a), whereas it indicates low power density in (b) to make object boundaries easier to distinguish.}
    \label{fig:power_density}
\end{figure}

P3DP addresses this problem using power density. Rather than measuring only the power of a particular cell, power density indicates how densely cells with similar power values are distributed around it. As illustrated in Fig.~\ref{fig:power_density}(a), a region containing many similar power responses has high power density. In contrast, power changes sharply near an object's surface or boundary, producing low power density. Similar to the KDE concept used in SMURF \cite{liu2023smurf,chen2017tutorial}, the density increases when more similar values occur in a local neighborhood.

To extract power density effectively, P3DP applies a Difference of Gaussians (DoG) \cite{bundy1984dog}, which measures the difference between the responses of two Gaussian filters with different sizes. Applying these filters to the tensor-power distribution around a candidate point produces the density response illustrated in Fig.~\ref{fig:power_density}(b).

\subsubsection{Density Filtering}

Candidate points are selected on the basis of their estimated power density. Points with low power density are considered more likely to be associated with an object's surface or boundary and are retained in the final point cloud. Points with high power density are considered less relevant to object shape and are removed as noise. In the experiments, P3DP retains the bottom 50\% of candidate points according to power density. This procedure produces a radar representation that suppresses more noise than simple percentile-based extraction while effectively preserving the object's 3D shape.

\subsection{4D Radar Point-Cloud Preprocessing: MF-KDE}

\begin{figure}[!htbp]
    \centering
    \includegraphics[width=\columnwidth]{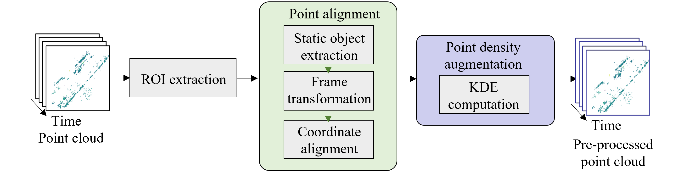}
    \caption{Overall pipeline of MF-KDE.}
    \label{fig:mfkde_pipeline}
\end{figure}

A point cloud contains few measurements per object because it is produced through detection and filtering. Its overall sparsity makes object shape difficult to represent and object-related points difficult to distinguish from noise. To address these limitations, we propose Multi-frame-based Noise Point Discrimination using Kernel Density Estimation (MF-KDE). MF-KDE increases the number of measurements through multi-frame alignment and adds a KDE-based point-density feature to improve the separability of object-related and noise points. As shown in Fig.~\ref{fig:mfkde_pipeline}, the procedure comprises ROI extraction, point alignment, and point-density augmentation.

\subsubsection{ROI Extraction}

As in P3DP, MF-KDE uses the ROI in \eqref{eq:roi} to retain only the spatial region needed for object detection in an autonomous driving environment.

\subsubsection{Point Alignment}

To address the sparsity of a radar point cloud, MF-KDE applies conventional frame alignment \cite{tan2022multiframe} to compensate for ego-vehicle motion. Points from several previous frames are transformed into the coordinate system of the current frame and accumulated into a single point cloud. This procedure supplements the object-shape information that is insufficient in a single frame. Simple accumulation, however, also increases the effects of noise and alignment errors, motivating an additional feature that represents the reliability of the accumulated points.

\subsubsection{Point-Density Augmentation}

\begin{figure}[!htbp]
    \centering
    \includegraphics[width=\columnwidth]{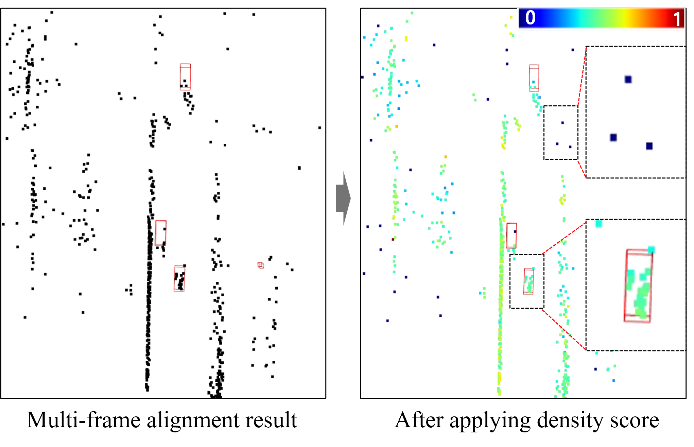}
    \caption{Visualization of point-density augmentation: (a) multi-frame alignment and (b) the result after adding the KDE-based point-density feature. Color represents the normalized density value (0: blue; 1: red). Points around objects exhibit relatively high density, whereas isolated points exhibit low density, providing an auxiliary feature for distinguishing object points from noise.}
    \label{fig:point_density}
\end{figure}

MF-KDE applies KDE to the multi-frame-aligned radar point cloud, calculates a local density value for each point, and uses it as a DNN input feature. KDE \cite{liu2023smurf,chen2017tutorial} places a kernel function around each point and estimates the density at that location from the spatial distribution of neighboring points. As shown in Fig.~\ref{fig:point_density}, points repeatedly observed from the same object form a spatially dense region after alignment, whereas points caused by noise or false alarms tend to have relatively low density. The KDE-based density therefore provides auxiliary information for distinguishing object-related points from noise.

Each point in a conventional radar point cloud is represented by its coordinates, Doppler velocity, RCS, and related radar information. MF-KDE adds the KDE-based density $\rho_i$ to construct the following feature vector:

\begin{equation}
    \mathbf{p}_{i}
    = \left[x_i, y_i, z_i, v_i, r_i, \rho_i\right],
    \label{eq:point_feature}
\end{equation}
where $x_i$, $y_i$, and $z_i$ are the 3D coordinates of point $i$; $v_i$ is its Doppler velocity; $r_i$ is its RCS or radar reflectivity; and $\rho_i$ is its KDE-based density.

\section{Embedded-System-Oriented Evaluation Metric}

Radar-only 3D object-detection models are generally evaluated using metrics such as AP, NDS, and frames per second (FPS). Because each of these metrics separately expresses detection accuracy, 3D detection quality, or processing speed, they cannot comprehensively assess suitability for an in-vehicle embedded perception system. Radar preprocessing can also affect the computational load and complexity of a DNN. We therefore propose Embedded \& NetScore (ENS), which jointly considers accuracy and efficiency:

\begin{equation}
    \Omega_N
    = 20\log\!\left(
    \frac{a_N^{\alpha} r_N^{\alpha} n_N^{\alpha}}
         {p_N^{\beta} m_N^{\gamma}}
    \right),
    \label{eq:ens}
\end{equation}
where $a_N$ is AP under normal driving conditions, $r_N$ is AP under adverse weather conditions, $n_N$ is NDS, $p_N$ is the number of network parameters, $m_N$ is the number of multiply-accumulate (MAC) operations, and $\alpha$, $\beta$, and $\gamma$ are weighting coefficients.

ENS builds on the efficiency-evaluation concept of NetScore, and each term reflects a requirement of an in-vehicle embedded perception system. The term $a_N$ represents the model's basic object-detection accuracy. The term $r_N$ quantifies robustness to adverse weather, a principal advantage of 4D radar. The term $n_N$ evaluates 3D object-detection quality by jointly considering errors in object range, size, orientation, velocity, and attributes rather than only whether an object is detected. The denominator contains efficiency terms related to embedded deployment: $p_N$ represents model size and memory demand, whereas $m_N$ represents the computation required for inference and, consequently, the potential for real-time processing. ENS is therefore designed to assign a high score to a radar-preprocessing and object-detection combination that achieves high accuracy and detection quality while operating efficiently under limited hardware resources.

The detection-accuracy and 3D-detection-quality terms $a_N$, $r_N$, and $n_N$ appear in the numerator, whereas the model-complexity and computational-cost terms $p_N$ and $m_N$ appear in the denominator. Thus, among models with similar detection performance, the model requiring fewer parameters and operations obtains a higher ENS. Following the design principles of NetScore \cite{wong2019netscore}, we set $\alpha=2$, $\beta=0.5$, and $\gamma=0.5$. NetScore gives accuracy a high weight because a model with low accuracy has limited practical utility even if it is small and computationally inexpensive \cite{wong2019netscore}. ENS extends this principle to 3D object detection by emphasizing accuracy under both normal and adverse weather and the quality of estimates of object position, size, orientation, and velocity, while penalizing parameter count and computational cost.

\section{Experiments}

This section uses the proposed ENS metric to evaluate the proposed 4D radar preprocessing methods and identify DNN models suitable for autonomous driving environments.

\subsection{Experimental Setup}

We describe the datasets, comparison methods, radar-based 3D object-detection models, and evaluation metrics used to validate P3DP and MF-KDE.

\subsubsection{Datasets}

We use K-Radar \cite{paek2022kradar} for the tensor-preprocessing experiments and Dual Radar \cite{zhang2025dual}---which includes Continental ARS548 RDI data---for the point-cloud-preprocessing experiments. K-Radar is the only publicly available dataset that provides a complete 4D radar tensor over the range, azimuth, elevation, and Doppler dimensions, making it suitable for validating 4D-tensor-preprocessing methods. However, K-Radar has a relatively narrow Doppler span, which can cause velocity overflow in real-world driving environments \cite{kong2025rtnh}. This limits its suitability for evaluating point-cloud preprocessing that actively uses Doppler information. We therefore use Dual Radar for the point-cloud experiments because it provides reliable Doppler information and very few points per frame, making the effect of the proposed method easier to evaluate. A total of five frames are aligned.

\subsubsection{Comparison Methods}

P3DP is compared with CA-CFAR (Cell-Averaging Constant False Alarm Rate) \cite{jalil2016analysis}, a representative baseline for point extraction from 4D radar tensors, and percentile-based extraction using the top 5\% of cells \cite{paek2022kradar}. MF-KDE is compared with conventional multi-frame alignment \cite{tan2022multiframe}, which transforms previous point clouds into the coordinate system of the current frame and accumulates them, and with DoppDrive \cite{haitman2025doppdrive}.

\subsubsection{Object-Detection Models}

P3DP is evaluated with RTNH \cite{paek2022kradar}, RadarPillar-Net \cite{musiat2024radarpillars}, and RPFA-Net \cite{xu2021rpfa}. RTNH uses a voxel-based object-detection architecture, whereas RadarPillar-Net and RPFA-Net use pillar-based architectures. Comparing models with different input representations and backbones evaluates the general effect of the proposed tensor-preprocessing method. MF-KDE is evaluated with PointPillars \cite{lang2019pointpillars}, RadarPillar-Net \cite{musiat2024radarpillars}, RPFA-Net \cite{xu2021rpfa}, and MF-Net \cite{tan2022multiframe}. MF-Net already uses multi-frame alignment and includes a spatiotemporal encoder for the aligned data, whereas PointPillars, RadarPillar-Net, and RPFA-Net use a single frame as their standard input.

\subsubsection{Evaluation Metrics}

We evaluate bird's-eye-view object-detection accuracy using $\mathrm{BEV}_{\mathrm{AP}}$, 3D bounding-box detection accuracy using $\mathrm{3D}_{\mathrm{AP}}$, and processing speed using FPS. Following the K-Radar benchmark, AP is calculated for the Sedan class at an intersection-over-union threshold of 0.3 \cite{paek2022kradar,jung2026lidar,kong2025rtnh}. For consistency, evaluation on Dual Radar is also restricted to the Car (Sedan) class. ENS is additionally used to consider accuracy, real-time performance, robustness, and model complexity jointly. Real-time performance is measured on an NVIDIA RTX 3090 GPU.

\subsection{Evaluation of 4D Radar Tensor Preprocessing}

\begin{figure}[!htbp]
    \centering
    \includegraphics[width=0.93\columnwidth]{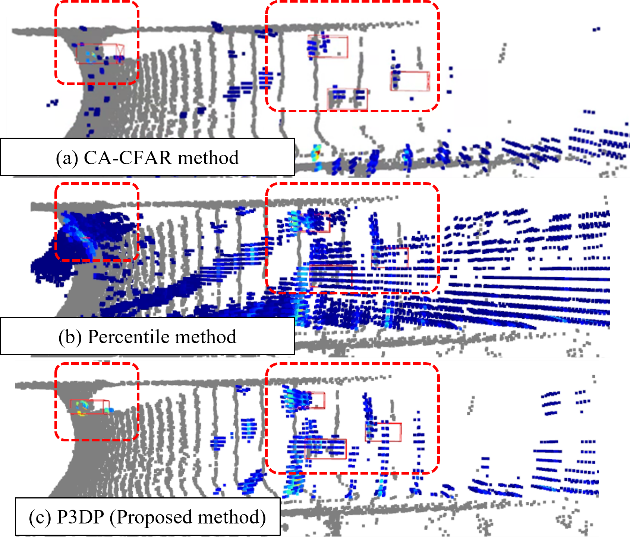}
    \caption{Comparison of tensor-preprocessing results.}
    \label{fig:tensor_results}
\end{figure}

Table~\ref{tab:tensor_preprocessing} and Fig.~\ref{fig:tensor_results} present the quantitative and qualitative results of 4D radar tensor preprocessing. P3DP generally achieves high $\mathrm{BEV}_{\mathrm{AP}}$, $\mathrm{3D}_{\mathrm{AP}}$, and ENS with RTNH, RadarPillar-Net, and RPFA-Net. This improvement is attributed to effective removal of unnecessary points while preserving object-shape information. P3DP adds DoG-based tensor power-density computation, which incurs an additional $O(N)$ cost in the number of tensor cells $N$ for fixed filter sizes and reduces FPS for the pillar-based detectors. For RTNH, however, P3DP reduces the number of active voxels and thereby reduces the cost of sparse 3D convolution, offsetting the additional preprocessing cost and increasing FPS. Parallel Gaussian filtering and preprocessing at the sensor or on dedicated hardware may further improve processing speed. Overall, RadarPillar-Net with P3DP achieves the highest ENS.

\begin{table}[!htbp]
    \centering
    \caption{Results of tensor preprocessing.}
    \label{tab:tensor_preprocessing}
    \renewcommand{\arraystretch}{1.10}
    \resizebox{\columnwidth}{!}{%
    \begin{tabular}{llcccc}
        \toprule
        \textbf{Detection model} & \textbf{Preprocessing} &
        $\mathbf{BEV_{AP}}$ [\%] & $\mathbf{3D_{AP}}$ [\%] &
        \textbf{FPS} [Hz] & \textbf{ENS} [\%] \\
        \midrule
        \multirow{3}{*}{RTNH \cite{paek2022kradar}}
        & CA-CFAR & 64.42 & 55.47 & 22.87 & 52.64 \\
        & Percentile 5\% & 64.89 & 56.05 & 22.65 & 53.84 \\
        & P3DP & \textbf{66.58} & \textbf{59.99} & \textbf{25.60} & \textbf{54.21} \\
        \midrule
        \multirow{3}{*}{RadarPillar-Net \cite{musiat2024radarpillars}}
        & CA-CFAR & 63.67 & 54.12 & 36.45 & 64.18 \\
        & Percentile 5\% & 57.63 & 54.10 & \textbf{36.56} & 63.49 \\
        & P3DP & \textbf{65.86} & \textbf{55.69} & 27.29 & \textbf{64.43} \\
        \midrule
        \multirow{3}{*}{RPFA-Net \cite{xu2021rpfa}}
        & CA-CFAR & 47.53 & 39.67 & 32.53 & 60.62 \\
        & Percentile 5\% & 48.59 & 46.05 & \textbf{35.50} & 61.58 \\
        & P3DP & \textbf{64.33} & \textbf{53.99} & 27.04 & \textbf{64.26} \\
        \bottomrule
    \end{tabular}}
\end{table}

For comparison with a recent tensor-based method, RADE-Net \cite{leitgeb2026rade} achieves a BEV AP of 68.7\% and a 3D AP of 64.1\% under the same K-Radar evaluation protocol, exceeding the best P3DP results by 3.18 and 6.85 percentage points, respectively. Its Tensor Projection Module, however, requires approximately 1.1~s per frame, which is burdensome for real-time processing. Thus, RADE-Net retains tensor information to obtain higher accuracy, whereas P3DP converts the tensor into a point-cloud representation to balance accuracy and real-time performance.

\subsection{Evaluation of 4D Radar Point-Cloud Preprocessing}

Table~\ref{tab:pointcloud_preprocessing} presents the results of 4D radar point-cloud preprocessing. MF-KDE improves $\mathrm{BEV}_{\mathrm{AP}}$, $\mathrm{3D}_{\mathrm{AP}}$, and ENS over the original input for most object-detection models. Multi-frame alignment supplements missing measurements, and the KDE-based density feature helps distinguish object-related points from noise. Accumulating multiple frames increases the number of input points, however, and density computation adds a $k$-d-tree-based neighbor search \cite{bentley1975multidimensional}, reducing FPS. For $M$ points and an average of $\overline{K}$ neighbors per point, density computation requires approximately $O(M\log M + M\overline{K})$ operations, and the larger number of points also increases the subsequent encoding and model computation. Parallelization and preprocessing at the sensor or on dedicated hardware may reduce this burden. Nevertheless, MF-KDE generally improves ENS and obtains the highest ENS when applied to MF-Net, demonstrating the benefit of using multi-frame information and density features spatiotemporally.

\begin{table}[!htbp]
    \centering
    \caption{Results of point-cloud preprocessing.}
    \label{tab:pointcloud_preprocessing}
    \renewcommand{\arraystretch}{1.05}
    \resizebox{\columnwidth}{!}{%
    \begin{tabular}{llcccc}
        \toprule
        \textbf{Detection model} & \textbf{Preprocessing} &
        $\mathbf{BEV_{AP}}$ [\%] & $\mathbf{3D_{AP}}$ [\%] &
        \textbf{FPS} [Hz] & \textbf{ENS} [\%] \\
        \midrule
        \multirow{4}{*}{PointPillars \cite{lang2019pointpillars}}
        & Not applied & 33.30 & 31.05 & \textbf{88.09} & 48.55 \\
        & Alignment & 33.60 & 32.57 & 33.89 & 48.76 \\
        & DoppDrive & 33.65 & 32.64 & 31.28 & 49.02 \\
        & MF-KDE & \textbf{33.68} & \textbf{32.72} & 29.57 & \textbf{49.35} \\
        \midrule
        \multirow{4}{*}{RadarPillar-Net \cite{musiat2024radarpillars}}
        & Not applied & 34.14 & 33.63 & \textbf{63.48} & 49.00 \\
        & Alignment & 34.29 & 33.85 & 28.56 & 49.08 \\
        & DoppDrive & 37.05 & \textbf{33.98} & 26.48 & 50.26 \\
        & MF-KDE & \textbf{38.17} & 32.94 & 25.15 & \textbf{50.92} \\
        \midrule
        \multirow{4}{*}{RPFA-Net \cite{xu2021rpfa}}
        & Not applied & 26.97 & 25.99 & \textbf{65.05} & 46.45 \\
        & Alignment & 31.79 & 31.59 & 25.31 & 48.31 \\
        & DoppDrive & 32.54 & 31.95 & 24.47 & 48.59 \\
        & MF-KDE & \textbf{32.92} & \textbf{32.14} & 25.19 & \textbf{50.03} \\
        \midrule
        \multirow{3}{*}{MF-Net \cite{tan2022multiframe}}
        & Alignment & \textbf{40.51} & 33.77 & \textbf{25.31} & 53.15 \\
        & DoppDrive & 40.44 & 34.84 & 23.28 & 53.10 \\
        & MF-KDE & 40.16 & \textbf{39.21} & 21.29 & \textbf{54.17} \\
        \bottomrule
    \end{tabular}}
\end{table}

We also compare MF-KDE with DoppDrive \cite{haitman2025doppdrive}, a recent point-cloud-preprocessing method. DoppDrive uses Doppler velocity to correct historical points from moving objects in the radial direction, reducing their dispersion during multi-frame accumulation, and achieves higher FPS than MF-KDE. However, it does not directly correct tangential motion and limits the accumulation duration of historical points with large expected errors, potentially reducing spatiotemporal information. MF-KDE instead retains the aligned multi-frame points and adds a KDE-based density feature that helps distinguish object points from noise, resulting in generally higher detection performance.

\section{Conclusion}

This paper proposed tensor- and point-cloud-preprocessing methods for 4D-radar-based 3D object detection. P3DP suppresses noise by extracting a point cloud from a 4D radar tensor while preserving the 3D shape of objects. MF-KDE improves the density and reliability of sparse point clouds through multi-frame alignment and a KDE-based density feature. We also proposed ENS, an evaluation metric that prioritizes detection accuracy and 3D detection quality while considering real-time performance and model complexity. Experimental results show that the proposed preprocessing methods generally improve both 4D-radar-based 3D object-detection performance and ENS. For deployment in a vehicle, candidate models can first be screened according to minimum detection-performance and real-time-processing requirements and available hardware resources; among the remaining models, the combination with the highest ENS can then be selected to account for the tradeoff between accuracy and real-time performance.

\section*{Acknowledgment}

This work was supported by Hyundai Motor Company and the Ministry of SMEs and Startups, Republic of Korea, under Grant RS-2025-24535910.

\bibliographystyle{IEEEtran}
\bibliography{references}

\end{document}